\documentclass[runningheads]{llncs}

\usepackage{eccv}
\usepackage{eccvabbrv}
\usepackage{graphicx}
\usepackage{booktabs}
\usepackage{tabularx}
\usepackage{array}
\usepackage{amsmath,amssymb,mathtools}
\usepackage{xcolor}
\usepackage{xspace}
\usepackage{microtype}
\usepackage{float}
\usepackage{marvosym}
\usepackage[section]{placeins}
\usepackage[accsupp]{axessibility}
\usepackage[breaklinks,colorlinks,citecolor=eccvblue,linkcolor=eccvblue,urlcolor=eccvblue]{hyperref}

\makeatletter
\renewcommand{\@fnsymbol}[1]{\ensuremath{\ifcase#1\or\star\or\text{\Letter}\or
  {\star\star}\or \dagger\or \ddagger\or
  \mathchar "278\or \mathchar "27B\or \|\or **\or \dagger\dagger
  \or \ddagger\ddagger \else\@ctrerr\fi}}
\makeatother

\newcommand{\JF}{\mathcal{J}\&\mathcal{F}}
\newcommand{\Fdot}{\dot{\mathcal{F}}}
\newcommand{\JFdot}{\mathcal{J}\&\dot{\mathcal{F}}}
\newcommand{\Nacc}{\mathrm{N\mbox{-}acc.}}
\newcommand{\Tacc}{\mathrm{T\mbox{-}acc.}}

\newcommand{\teaminfo}[3]{%
  \vspace{0.2mm}
  \begin{center}
  \renewcommand{\arraystretch}{0.92}%
  \setlength{\tabcolsep}{3pt}%
  \begin{tabularx}{\linewidth}{@{}>{\bfseries}lX@{}}
    \toprule
    Method & #1 \\
    Authors & #2 \\
    Affiliations & #3 \\
    \bottomrule
  \end{tabularx}
  \end{center}
  \vspace{0.2mm}
}

\title{Report of the 8th LSVOS Challenge:\
Complex and Multimodal Video Object Segmentation}
\titlerunning{Report of the 8th LSVOS Challenge}

\author{Chang Liu\thanks{Organizers of the 8th LSVOS Challenge, ECCV 2026. Following authors are the top-3 team members of each track.}, \quad
Henghui Ding\textsuperscript{*}\thanks{Corresponding to Henghui Ding (\email{henghui.ding@gmail.com}), the Institute of Big Data, Fudan University, Shanghai, China.}, \quad
Lingyi Hong\textsuperscript{*}, \quad
Ning Xu\textsuperscript{*}, Linjie Yang\textsuperscript{*}, \quad Yuchen Fan\textsuperscript{*}\\
\vspace{0.3cm}
Canyang Wu, \quad Jinrong Zhang, \quad Xusheng He, \quad Ce Bian, \quad Xianjing Han, \quad Jianlong Wu, \quad
Mingqi Gao, \quad Sijie Li, \quad Jungong Han, \quad
JeongRae Kim, \quad Chaehyun Kim, \quad Changwon Lim, \quad
Jungyoon Lee, \quad Gyuil Lim, \quad Doeon Kim, \quad Seong-heum Kim, \quad
Pranjal Aggarwal, \quad Sean Welleck, \quad Yiwen Ren,\\ \quad Jianing Liu, \quad Yingxin Wang, \quad Kexin Zhang, \quad Licheng Jiao, \quad Lingling Li, \quad Xu Liu, \quad
Jinxing Zhou, \quad Suiyi Zhao, \quad Yanghao Zhou, \quad Ruohao Guo, \quad
Liangtao Shi, \quad Jinxia Xie, \quad Xiantao Hu, \quad Ting Liu}
\authorrunning{Liu et al.}
\institute{}

\begin{document}
\maketitle

\begingroup
\centering
{\url{https://lsvos.github.io/}\par}
\endgroup

\begin{abstract}
This report summarizes the 8th Large-scale Video Object Segmentation (LSVOS) Challenge, held in conjunction with ECCV 2026. The challenge evaluates video segmentation in three complementary settings: complex semi-supervised video object segmentation on MOSEv2, text-guided referring video object segmentation on MeViSv2-Text, and audio-guided referring video object segmentation on MeViSv2-Audio. We describe the tasks and evaluation protocols and review the methods of the top three teams in each track. Across the nine leading solutions, foundation segmentation models are combined with target-aware memory, multimodal reasoning, explicit target-existence verification, agentic interaction, and corrective tracking. These systems illustrate a broader transition from single-model mask propagation toward modular pipelines that reason about object identity, query validity, and temporal reliability.
\end{abstract}

\section{Introduction}
\label{sec:overall-intro}

Video object segmentation (VOS) aims to delineate and track target objects throughout a video. Although modern foundation models have substantially improved mask quality and generalization, complex scenes remain difficult because targets may be small, heavily occluded, visually similar to distractors, or absent for long periods before reappearing. Benchmarks such as MOSE and MOSEv2 emphasize these failure modes and provide a demanding setting for studying reliable long-term target propagation~\cite{ding2023mose,ding2025mosev2,gres}.

Referring video object segmentation (RVOS) replaces the first-frame mask with a natural-language description of the target. MeViS introduced motion expressions as the principal cue for identifying objects, requiring a model to understand not only appearance but also actions, trajectories, temporal order, and interactions~\cite{ding2023mevis,liu2024primitivenet}. MeViSv2 extends this setting with more challenging expressions, including queries for which no valid target exists, and supports both text and spoken motion descriptions~\cite{ding2025mevisv2}. The resulting tasks require semantic grounding, temporal reasoning, dense mask prediction, and control of false-positive outputs.

The tracks are complementary. MOSEv2 tests identity preservation under interrupted or ambiguous visual evidence; MeViSv2-Text tests grounding from motion and temporal relations rather than appearance; and MeViSv2-Audio adds spoken-input uncertainty. Each requires semantic understanding, target-existence decisions, mask initialization, and long-term propagation, enabling comparison of VOS and multimodal-grounding strategies in a common framework.

The 8th LSVOS Challenge is held in conjunction with ECCV 2026 in Malm\"o, Sweden. We organize three tracks: MOSEv2, MeViSv2-Text, and MeViSv2-Audio~\cite{lsvos2026}. Together, these tracks cover mask-initialized VOS, text-guided RVOS, and audio-guided RVOS.

Beyond ranking submissions, we aim to document which designs remain reliable across these uncertainties. We describe the tracks, protocols, and official results, then summarize the top three solutions from their technical reports. Finally, we compare recurring components: foundation models, multimodal reasoning, memory, verification, and corrective re-initialization, and discuss emerging directions.

\section{The 8th LSVOS Challenge}
\label{sec:challenge-overview}

\subsection{Challenge Tracks}

\noindent\textbf{Track 1: Complex Video Object Segmentation (MOSEv2).}
Given a video and the target masks in the first frame, a method must segment the same object instances in all subsequent frames. MOSEv2 focuses on complex environments containing disappearance and reappearance, crowded scenes, heavy occlusion, small or inconspicuous targets, adverse capture conditions, and strong same-category distractors~\cite{ding2025mosev2,ding2023mose,ding2026grex}.

\noindent\textbf{Track 2: Text-based Referring Motion Expression Video Segmentation (MeViSv2-Text).}
Given a video and a textual motion expression, a method must identify every object satisfying the expression and predict its masks over time. The description may depend on motion, temporal composition, object interactions, or semantic roles rather than static category and appearance alone~\cite{ding2025mevisv2,ding2023mevis}.

\noindent\textbf{Track 3: Audio-based Referring Motion Expression Video Segmentation (MeViSv2-Audio).}
This track uses a spoken motion expression in place of text. A method must first interpret the audio and then ground the described target in the video. The task introduces transcription uncertainty while retaining the target-existence and mask-prediction requirements of MeViSv2~\cite{ding2025mevisv2,ding2023mevis}.

\subsection{Evaluation Protocol}

The primary ranking metric for MOSEv2 is $\JFdot$, the mean of region similarity $\mathcal{J}$ and adaptive boundary accuracy $\Fdot$~\cite{ding2025mosev2}. Unlike the classical boundary measure $\mathcal{F}$, $\Fdot$ adapts its boundary tolerance to object scale. MOSEv2 also reports $\JFdot_{\mathrm{d}}$ and $\JFdot_{\mathrm{r}}$ on disappearance and reappearance clips, respectively. The two MeViSv2 tracks instead use the classical $\JF$ together with target-existence accuracy. Their primary score combines mask quality, no-target accuracy, and target accuracy as $\mathrm{Final}=(\JF+\Nacc+\Tacc)/3$.
This protocol rewards accurate masks while explicitly penalizing systems that hallucinate a target for an invalid query or suppress a valid target.

\subsection{Challenge Results}

HITsz-Dragon leads MOSEv2 at 69.82 $\JFdot$; SSUPER and AEXBY lead the text and audio tracks with Final scores of 90.81 and 76.96, respectively.

Besides the ranked submissions, organizer-developed FudanSeg serves as an auxiliary MOSEv2 reference. It reaches 64.71 $\JFdot$, exceeding the SAM~3 baseline by 1.99 points, with gains of 1.89 in $\mathcal{J}$ and 2.10 in $\Fdot$. Its disappearance score rises from 78.85 to 84.75, while reappearance decreases from 25.32 to 23.65, indicating gains concentrated in disappearance handling and boundary quality. As an organizer method, it is excluded from the official ranking.

\begin{table}[H]
  \caption{MOSEv2 final test-set results (\%). Bold indicates the best ranked score; FudanSeg is an organizer entry and does not participate in the ranking.}
  \label{tab:mosev2-results}
  \centering
  \scriptsize
  \setlength{\tabcolsep}{8pt}%
  \renewcommand{\arraystretch}{0.95}%
  \begin{tabular}{@{}lrrr@{}}
    \toprule
    Team & $\JFdot$ & $\mathcal{J}$ & $\Fdot$ \\
    \midrule
    HITsz-Dragon & \textbf{69.82} & \textbf{68.21} & \textbf{71.43} \\
    mmm          & 66.20 & 64.79 & 67.60 \\
    AISTAT       & 64.37 & 63.16 & 65.59 \\
    \midrule
    FudanSeg$^\ast$ & 64.71 & 63.51 & 65.92 \\
    Baseline (SAM~3) & 62.72 & 61.62 & 63.82 \\
    \bottomrule
  \end{tabular}
\end{table}

\begin{table}[H]
  \caption{MeViSv2 final test-set leaderboard results (\%). Bold indicates the best score in each track.}
  \label{tab:mevisv2-results}
  \centering
  \scriptsize
  \setlength{\tabcolsep}{2.5pt}%
  \renewcommand{\arraystretch}{0.95}%
  \begin{minipage}[t]{0.49\linewidth}
    \centering
    \textbf{MeViSv2-Text}\par\smallskip
    \begin{tabular}{@{}lrrrr@{}}
      \toprule
      Team & $\JF$ & $\Nacc$ & $\Tacc$ & Final \\
      \midrule
      SSUPER          & \textbf{79.22} & \textbf{94.44} & \textbf{98.77} & \textbf{90.81} \\
      CUA Generalists & 74.96 & 88.89 & 96.32 & 86.72 \\
      HITsz-Dragon    & 76.10 & 83.33 & 97.55 & 85.66 \\
      \midrule
      Baseline & 41.26 & 100.00 & 65.69 & 68.98 \\
      \bottomrule
    \end{tabular}
  \end{minipage}\hfill
  \begin{minipage}[t]{0.49\linewidth}
    \centering
    \textbf{MeViSv2-Audio}\par\smallskip
    \begin{tabular}{@{}lrrrr@{}}
      \toprule
      Team & $\JF$ & $\Nacc$ & $\Tacc$ & Final \\
      \midrule
      AEXBY       & \textbf{59.52} & 79.31 & \textbf{92.05} & \textbf{76.96} \\
      StopTheRoll & 53.74 & 72.41 & 85.54 & 70.57 \\
      Agent-VOS   & 46.03 & \textbf{96.55} & 57.83 & 66.80 \\
      \midrule
      Baseline & 44.88 & 0.00 & 100.00 & 48.29 \\
      \bottomrule
    \end{tabular}
  \end{minipage}
\end{table}

\section{Top Solutions in the MOSEv2 Track}
\label{sec:mosev2-solutions}

\begin{figure}[t]
  \centering
  \begin{subfigure}[t]{0.485\linewidth}
    \centering
    \includegraphics[width=\linewidth]{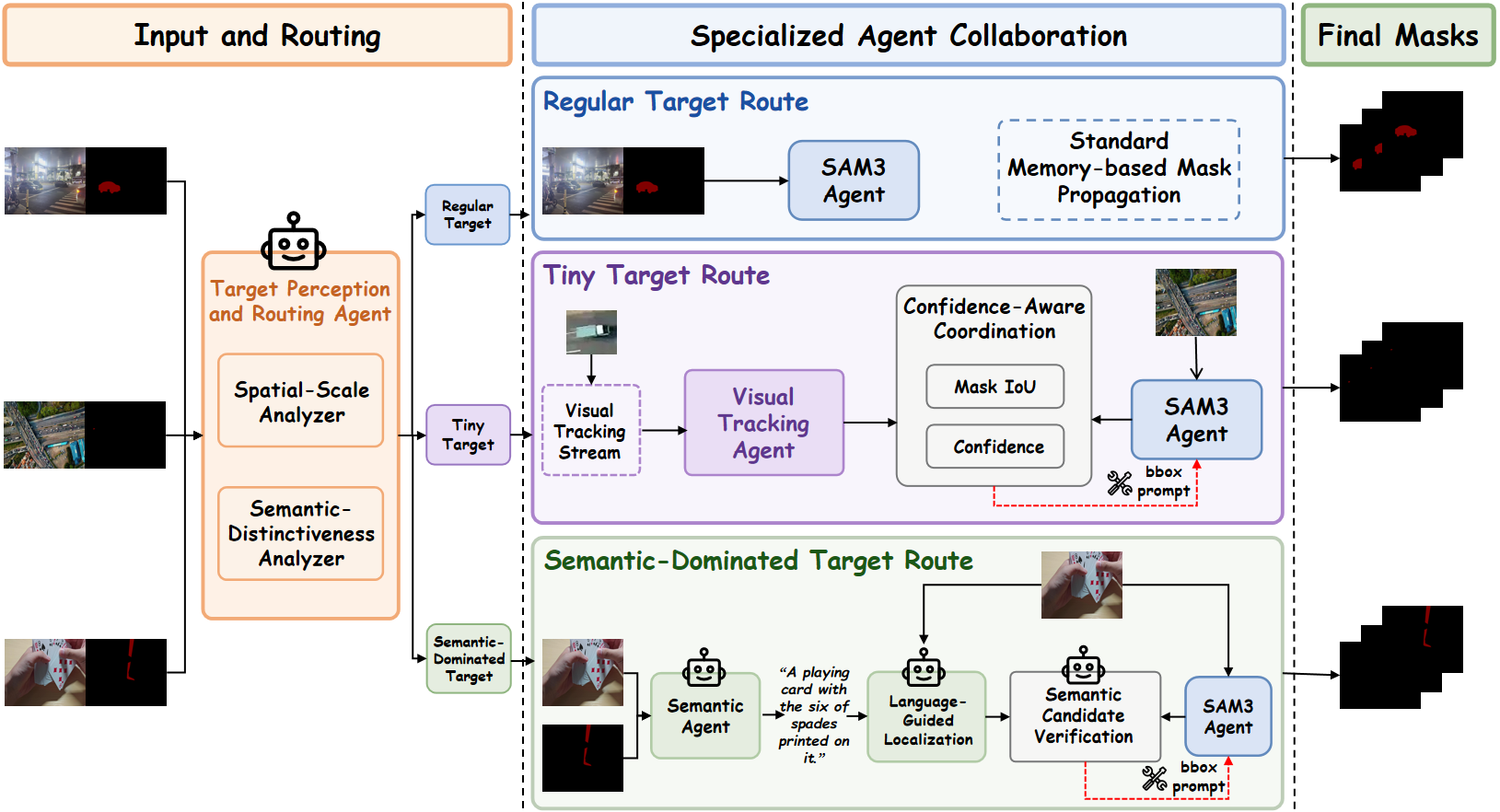}
    \caption{VOS-Agent routes regular, tiny, and semantic-dominated targets to specialized agents.}
    \label{fig:mose1-overview}
  \end{subfigure}\hfill
  \begin{subfigure}[t]{0.485\linewidth}
    \centering
    \includegraphics[width=0.92\linewidth]{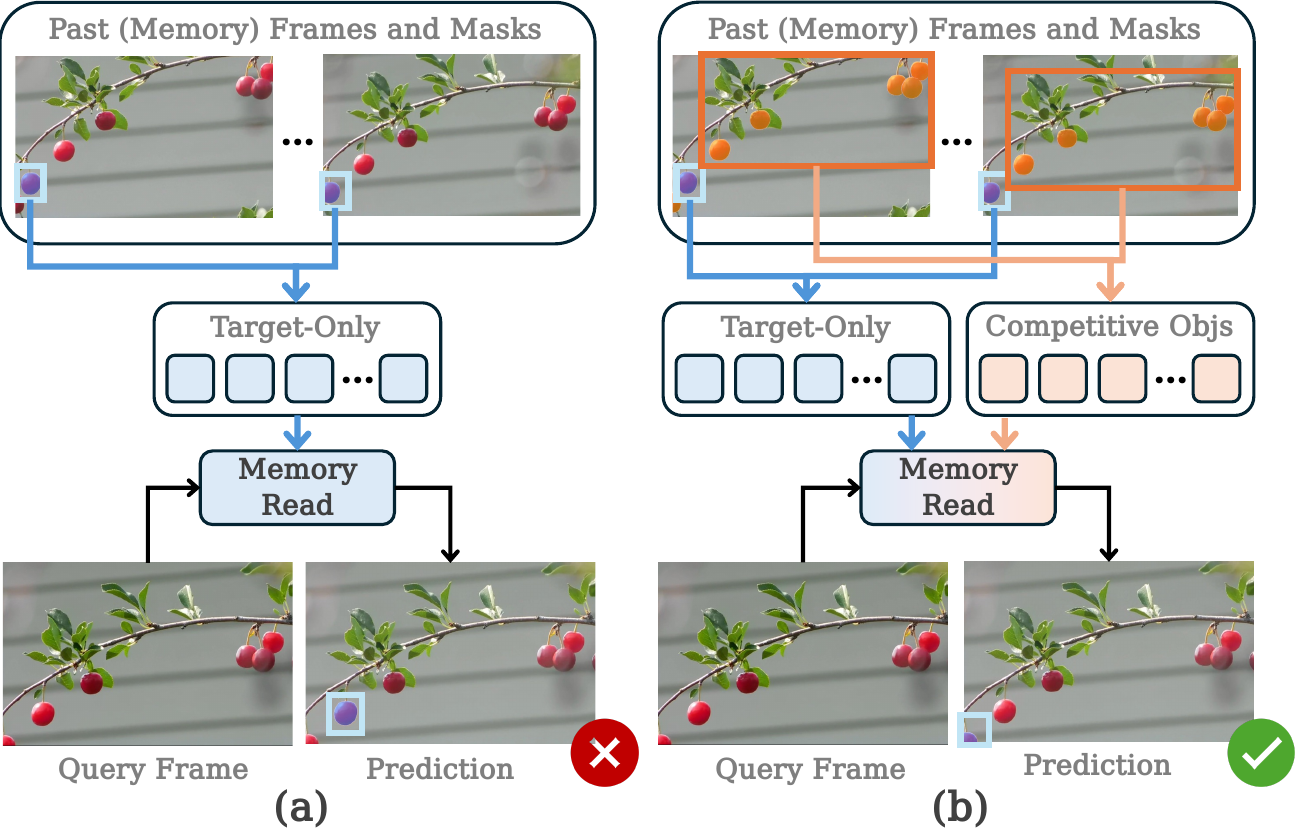}
    \caption{Competitive Memory Readout contrasts target evidence with same-class competitors.}
    \label{fig:mose2-overview}
  \end{subfigure}
  \caption{Method overviews of the first- and second-place MOSEv2 teams.}
  \label{fig:mosev2-method-overviews}
\end{figure}

\subsection{1st Place: HITsz-Dragon}
\label{sec:mose1}

\teaminfo{VOS-Agent}{Canyang Wu$^{1}$, Jinrong Zhang$^{1}$, Xusheng He$^{1}$, Ce Bian$^{1}$, Xianjing Han$^{2}$, Jianlong Wu$^{1,3}$}{$^{1}$Harbin Institute of Technology, Shenzhen, China; $^{2}$Nanyang Technological University, Singapore; $^{3}$Shenzhen Loop Area Institute, China}

Given a video $\mathcal{V}=\{I_t\}_{t=1}^{T}$ and the first-frame mask $M_1$ of a target, VOS-Agent predicts masks $\{\widehat{M}_t\}_{t=2}^{T}$ while adapting its inference route to the target's failure mode. As shown in Fig.~\ref{fig:mose1-overview}, the system contains a Target Perception and Routing Agent, a SAM~3 Segmentation Agent, a Visual Tracking Agent, and a Semantic Agent. Regular targets are handled by standard SAM~3 propagation; tiny targets receive corrective tracking prompts; and semantic-dominated targets receive identity-level verification.

\paragraph{Target perception and routing.}
The routing agent examines target scale and semantic distinctiveness. Let $A(M_1)$ and $A(I_1)$ denote the target and image areas. The normalized area $r_{\mathrm{area}}=A(M_1)/A(I_1)$ identifies tiny targets when $r_{\mathrm{area}}<\tau_{\mathrm{area}}$. For a non-tiny target, an MLLM examines the reference crop and context to determine whether explicit attributes---such as text, logos, symbols, color, or distinctive clothing---are necessary to preserve identity. The final route is regular unless either the scale test or semantic-distinctiveness test activates a specialized agent.

\paragraph{Shared SAM~3 segmentation agent.}
SAM~3 serves as the common dense segmentation and temporal propagation module~\cite{carion2026sam}. The initial mask creates the target masklet and conditioning memory. For every subsequent frame, the tracker combines current-frame features with the first-frame reference and confidently tracked history. A specialized agent may add a box prompt on a selected frame, after which SAM~3 refines the current mask and updates its conditioning state.

\paragraph{Tracking-agent collaboration.}
Tiny targets often provide too little pixel evidence for stable propagation. The visual tracking route uses SUTrack~\cite{chen2025sutrack}, initialized by the box enclosing $M_1$, to recursively estimate a target box $B_t^{\mathrm{trk}}$ and confidence $c_t^{\mathrm{trk}}$. In parallel, SAM~3 produces a mask box $B_t^{\mathrm{sam}}$. Their agreement is $q_t^{\mathrm{trk}}=\operatorname{IoU}(B_t^{\mathrm{sam}},B_t^{\mathrm{trk}})$. A tracking prompt is accepted only when the estimates disagree and the tracker is confident, namely $\eta_t^{\mathrm{trk}}=\mathbb{I}[q_t^{\mathrm{trk}}<\tau_{\mathrm{iou}}\land c_t^{\mathrm{trk}}\geq\tau_{\mathrm{conf}}]$.
When $\eta_t^{\mathrm{trk}}=1$, the tracking box corrects the current SAM~3 masklet; otherwise the native prediction is retained. The two agents therefore maintain complementary states: SUTrack estimates location through recursive templates, while SAM~3 maintains dense video memory.

\paragraph{Semantic-agent collaboration.}
For semantic-dominated targets, an MLLM first generates a discriminative description $D$ from the initial target crop. It then performs description-guided localization in later frames, producing $B_t^{\mathrm{sem}}$. If this box agrees with the SAM~3 box, the native mask is retained. When agreement is low, the semantic agent compares the initial target crop with both current-frame candidates and chooses the candidate that better preserves the designated identity. A semantic box is injected only when the language-guided candidate wins this explicit verification. In this way, auxiliary agents intervene selectively rather than replacing the shared segmentation backbone.

\subsection{2nd Place: mmm}
\label{sec:mose2}

\teaminfo{Competitive Memory Readout for Robust VOS}{Mingqi Gao$^{1}$, Sijie Li$^{1}$, Jungong Han$^{2}$}{$^{1}$School of Computer Science, University of Sheffield; $^{2}$Department of Automation, Tsinghua University}

The method retains the standard SAM~3 video segmentation backbone, memory encoder, and mask decoder, but changes how stored memory is read. Its central component, Competitive Memory Readout (CMR), augments target-only retrieval with same-class non-target competitors. A region may be highly similar to the target history while still belonging to the wrong instance; CMR therefore calibrates target evidence against plausible alternatives rather than treating all non-target regions as undifferentiated background. A deterministic adaptive-restoration rule complements this competition by recovering weak but correct targets after disappearance or long occlusion.

\paragraph{Competitive memory readout.}
For target object $o$, let $\mathcal{T}^{o}_{f}$ denote its foreground memory tokens in memory frame $f$. Same-class non-target hypotheses that do not strongly overlap the tracked target are encoded as competitor tokens $\mathcal{D}^{o}_{f}$. For current-frame token $i$, target and competitor evidence is aggregated as $r^{T}_{i,f}=\operatorname{LSE}_{u\in\mathcal{T}^{o}_{f}}\ell^{f}_{i,u}$ and $r^{C}_{i,f}=\operatorname{LSE}_{u\in\mathcal{D}^{o}_{f}}\ell^{f}_{i,u}$. The gate $g_{i,f}=\sigma((r^{T}_{i,f}-r^{C}_{i,f})/\tau_c)$ is applied to target foreground logits before the original memory-attention normalization. Target evidence is preserved when the query is better explained by the target memory and suppressed when a same-class competitor is stronger. If no valid competitor exists, the method reduces to the original target-only readout.

\paragraph{Adaptive restoration.}
Competition can suppress useful evidence when the true target is weak. Let $m^{f}_{\mathrm{pre}}$ and $m^{f}_{\mathrm{post}}$ be the response before and after competition. The relative suppression and restoration factor are
\begin{equation}
  s_f=\operatorname{clip}\!\left(
  \frac{m^{f}_{\mathrm{pre}}-m^{f}_{\mathrm{post}}}
       {m^{f}_{\mathrm{pre}}+\epsilon},0,1\right),
  \qquad \rho_f=1+0.5\sqrt{s_f}.
  \label{eq:mose2-restoration}
\end{equation}
The factor ranges from $1.0$ to $1.5$: weak competition receives little correction, while heavily suppressed target evidence receives stronger restoration. CMR and restoration thus balance identity discrimination and recoverability without an additional learned policy.

\subsection{3rd Place: AISTAT}
\label{sec:mose3}

\teaminfo{SAM3Dual}{JeongRae Kim, Chaehyun Kim, Changwon Lim}{Chung-Ang University, Seoul, Republic of Korea}

SAM3Dual extends pretrained SAM~3 with a training-free dual-memory mechanism for long-term VOS. All pretrained parameters remain frozen. The method separates temporal information into a short-term branch containing recent observations and a long-term branch containing temporally dispersed history. Current-frame features retrieve information independently from both branches with the same frozen memory-attention operation. The two responses are confidence-modulated and combined by a deterministic sequence-relative schedule.

\paragraph{Dual-memory architecture.}
Let $Q_t$ be the current-frame query representation. SAM3Dual maintains short- and long-term banks $\mathcal{M}^{S}_t$ and $\mathcal{M}^{L}_t$. Both permanently retain the first-frame ground-truth representation and up to six additional temporal entries. The short-term bank stores recent observations, whereas the long-term bank selects history at fixed intervals. Their responses, $O^{S}_t=\operatorname{CrossAttn}_{\theta}(Q_t,K^{S}_t,V^{S}_t)$ and $O^{L}_t=\operatorname{CrossAttn}_{\theta}(Q_t,K^{L}_t,V^{L}_t)$, use the same frozen parameters $\theta$ and differ only in temporal composition.

\paragraph{Confidence-guided modulation.}
The previous-frame object-confidence logit gives $c_{t-1}=\sigma(z_{t-1})$ and the conservative scale $s_t=s_{\min}+(s_{\max}-s_{\min})c_{t-1}$, with $s_{\min}=0.9$ and $s_{\max}=1.1$. Both memory responses are multiplied by $s_t$, slightly amplifying memory when confidence is high and attenuating it when confidence is low without changing the branches' relative weight.

\paragraph{Sequence-relative temporal fusion.}
For a sequence of $N$ frames, $g_t=1-(1-\alpha)t/(N-1)$ and $O_t=g_t\widetilde{O}^{S}_t+(1-g_t)\widetilde{O}^{L}_t$. The submitted system uses $\alpha=0.5$, so the short-term weight decreases from $1.0$ to $0.5$ as the long-term contribution increases. Normalizing by sequence length gives the same schedule to videos of different duration.

After each prediction, SAM~3 encodes the current mask into memory. The short-term branch retains recent entries, while the long-term branch samples history every $\Delta=10$ frames. Both remain bounded and retain the first-frame reference. The image encoder, memory encoder, attention modules, and mask decoder are never updated, so the complete method requires no task-specific training or online optimization.

\FloatBarrier

\section{Top Solutions in the MeViSv2-Text Track}
\label{sec:text-solutions}

\begin{figure}[t]
  \centering
  \begin{subfigure}[t]{0.49\linewidth}
    \centering
    \includegraphics[width=\linewidth,trim=0 80bp 0 0,clip]{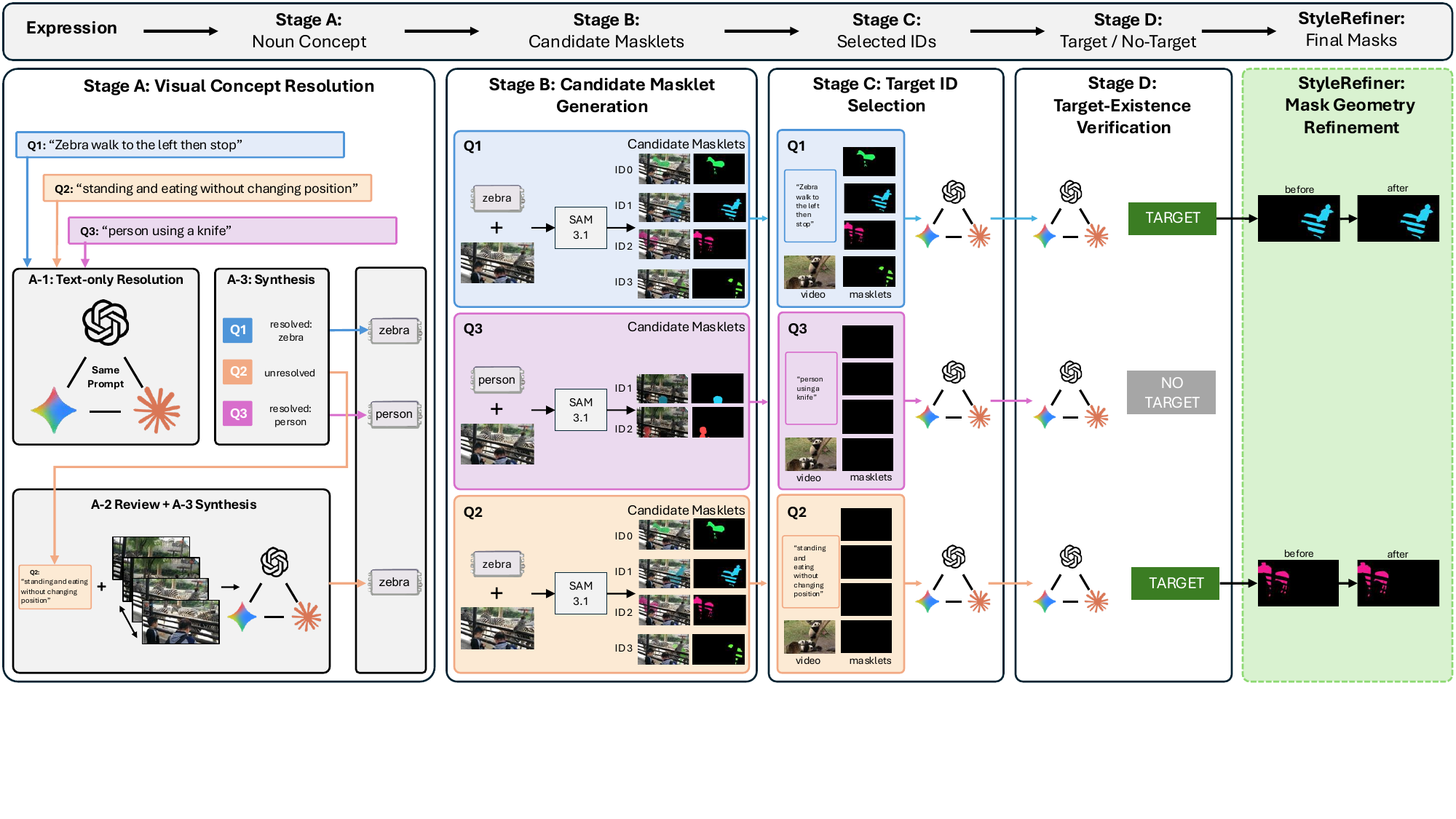}
    \caption{SSUPER performs multi-agent grounding, existence verification, and mask refinement.}
    \label{fig:text1-overview}
  \end{subfigure}\hfill
  \begin{subfigure}[t]{0.49\linewidth}
    \centering
    \includegraphics[width=\linewidth]{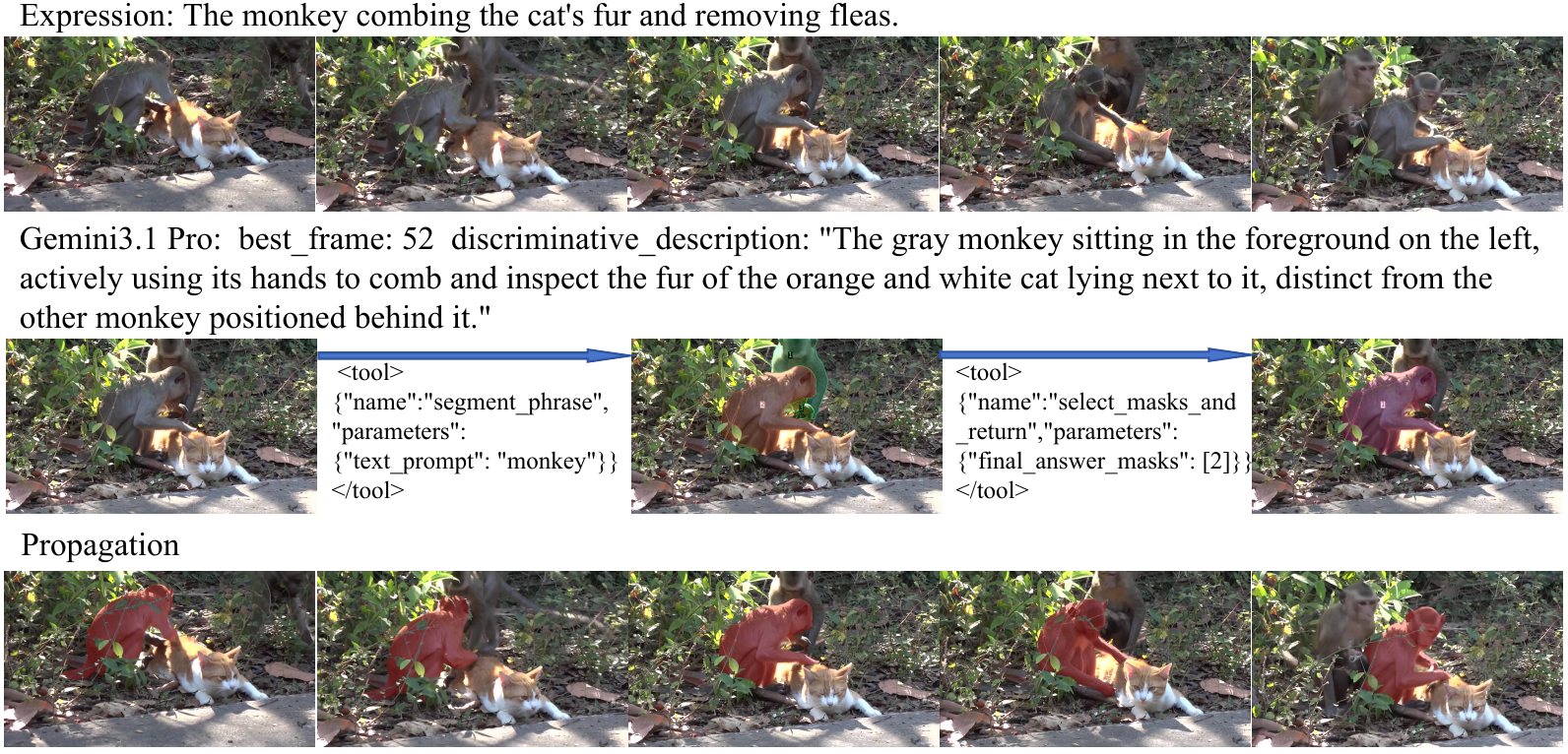}
    \caption{HITsz-Dragon converts an event into instance descriptions and seed masks.}
    \label{fig:text3-pipeline}
  \end{subfigure}
  \caption{Representative pipelines from the MeViSv2-Text track.}
  \label{fig:text-method-overviews}
\end{figure}

\subsection{1st Place: SSUPER}
\label{sec:text1}

\teaminfo{Multi-Agent Target-Existence Verification and Learned Mask Geometry Refinement}{Jungyoon Lee$^{1}$, Gyuil Lim$^{2}$, Doeon Kim$^{3}$, Seong-heum Kim$^{1,2,3}$}{$^{1}$Department of AI Convergence Security, $^{2}$Department of AI Convergence, and $^{3}$Department of Intelligent Semiconductors, Soongsil University, Republic of Korea}

SSUPER uses a track-before-selection design with four semantic stages followed by a learned geometry refiner. Given a video and motion expression, the system first converts the expression into a visual concept, asks SAM~3.1 to generate candidate masklets, selects candidates by comparing their complete temporal behavior with the original expression, and then performs a separate audit of target existence. The final StyleRefiner changes mask geometry only after all semantic decisions have been fixed.

\paragraph{Shared multi-agent protocol.}
Stages A, C, and D use the same review--synthesis block. Three heterogeneous MLLMs receive an identical stage-specific prompt and identical ordered visual evidence, then independently return schema-validated outputs. A synthesis call receives the evidence and all three responses in fixed order and commits one result under the same schema. It may resolve disagreement, but cannot introduce a candidate identity that was not supplied by the mask generator. Contact sheets preserve chronology, exact frame names, and consistent numbering so that agents can cite the evidence used for each decision.

\paragraph{Track-first grounding.}
Stage A extracts an expression-wise visual concept. A text-only review first tries to produce a short noun phrase suitable for SAM~3.1. If the expression is noun-less or visually dependent, the same review is repeated with ordered video frames. Action, direction, temporal order, count, and relations remain in the original expression for later selection.

In Stage B, the visual concept prompts the SAM~3.1 video predictor with Object Multiplex~\cite{meta2026sam31}. The predictor returns full-video masklets with stable IDs, masks, boxes, and confidence. It does not decide which candidate satisfies the referring expression. Late entry and occlusion are represented as frame-level empty regions inside a full-video candidate rather than being mistaken for expression-level absence.

Stage C receives the original expression, chronological RGB frames, and numbered masklet overlays. Each review agent compares candidate appearance and temporal behavior and returns one or more IDs, \texttt{no\_target}, or \texttt{unresolved}; the synthesis pass commits the result. The system also archives the best non-empty candidate set before existence gating, allowing the following stage to distinguish a semantic rejection from the absence of any plausible track.

\paragraph{Decoupled existence verification.}
Stage D asks whether any object in the complete video satisfies the \emph{full} predicate. It independently audits category and appearance, count, action or state, direction and trajectory, temporal composition, relations, and actor--patient roles. Reviews span the beginning, middle, and end of the video, distinguish temporary invisibility from true absence, and discount apparent motion caused by camera movement. A no-target verdict requires evidence contradicting at least one required predicate; uncertainty alone is not converted into absence.

Stage D generates no geometry. A no-target verdict exports an empty sequence, while a target verdict preserves a non-empty Stage C result. If Stage C exported empty masks, Stage D may restore only the same expression's archived provisional IDs. It cannot call SAM~3.1, create candidates, or borrow an identity from another expression. This separation isolates target-existence reasoning from proposal generation.

\paragraph{StyleRefiner.}
After the presence decision is fixed, StyleRefiner aligns SAM-derived masks with MeViSv2 annotation geometry. Training pairs are built from MeViSv2 training annotations and frozen SAM~3.1 predictions; videos, rather than frames, are split between training and development. A five-channel crop---normalized RGB, mask prior, and signed distance transform---is processed by a ConvNeXt-S encoder, a high-resolution detail stem, and a U-Net-like decoder~\cite{liu2022convnext,simeoni2025dinov3,ronneberger2015unet}. At inference, each sufficiently large connected component is refined independently and pasted back at the original resolution. Empty frames, small components, and erased predictions fall back to the input, and the Stage D presence decision is preserved as a final invariant.

\subsection{2nd Place: CUA Generalists}
\label{sec:text2}

\teaminfo{Multimodal Agents Using Task-Specific Software}{Pranjal Aggarwal, Sean Welleck}{Language Technologies Institute, Carnegie Mellon University}

This solution formulates MeViSv2-Text prediction as interaction between a general multimodal agent and task-specific software. The design reuses the pointing action familiar from computer-use agents: in a graphical application, a click selects the element on which software should operate; in structured vision prediction, a point selects the object on which segmentation and tracking tools should operate. The agent reasons about the video and expression, while the software owns mask generation, temporal propagation, state management, and serialization into the challenge format.

For an input $x$ and required structured output $y$, the software $S$ exposes a set of task actions $\mathcal{A}_S$ and a submission action. The agent receives $x$, action definitions, and the observations returned by $S$. Every action updates the software state and returns a new observation. On submission, the software serializes its state as $y$. The agent and interaction loop remain fixed, while the task determines the available actions and output format. This approach is related to computer-use agents that operate software through pixels, clicks, and typing~\cite{aggarwal2025pwp,aggarwal2026gymanything}, but exposes video segmentation and tracking as explicit, typed operations.

\paragraph{Actions and state.}
Each trajectory begins with the full source video and motion expression. The software provides six actions: \texttt{select\_object}, \texttt{refine\_object}, \texttt{remove\_object}, \texttt{preview\_tracking}, \texttt{submit}, and \texttt{submit\_no\_target}. To select an object, the agent specifies a frame and normalized image point. The software converts this location into a positive SAM~3 prompt and initializes a separate object identity. Multiple calls create multiple identities, whose masks are merged into the binary output required by the challenge.

The refinement action adds positive or negative points to an existing identity, while removal deletes an incorrect identity. When the agent requests a preview, SAM~3 propagates every identity through the video and returns a color-coded tracked-mask overlay. The agent can compare the overlay with the expression, correct an identity, add a missing object, remove a distractor, or request another preview. Because the complete source video remains available throughout the trajectory, the agent can revisit the evidence rather than relying only on its most recent view.

\paragraph{Prediction and export.}
The motion expression conditions the agent's selections and refinement points; SAM~3 converts those points into masks and propagates them. Submission preserves the original frame dimensions and writes one indexed PNG per required frame in the official directory structure. For an expression with no matching object, \texttt{submit\_no\_target} writes an empty mask sequence. The central contribution is therefore not a new mask backbone, but a reusable interface that lets a multimodal agent construct, inspect, and revise a structured video prediction before committing it.

\subsection{3rd Place: HITsz-Dragon}
\label{sec:text3}

\teaminfo{Event Decomposition and SAM3-Agent Propagation}{Ce Bian$^{1}$, Xusheng He$^{1}$, Jinrong Zhang$^{1}$, Canyang Wu$^{1}$, Xianjing Han$^{2}$, Jianlong Wu$^{1,3}$}{$^{1}$Harbin Institute of Technology, Shenzhen, China; $^{2}$Nanyang Technological University, Singapore; $^{3}$Shenzhen Loop Area Institute, China}

The method adopts a two-stage framework comprising event understanding, single-frame localization, and video propagation. An MLLM first analyzes the video and motion expression, decomposes the query into one or more instance-level targets, selects a localization key frame for each target, and produces a discriminative description conditioned on that frame. A SAM3-agent then generates a pixel-level seed mask, which the SAM~3 video tracker propagates in both temporal directions. Multiple valid instances are processed independently and merged into an event-level prediction.

\paragraph{Event decomposition and key-frame reasoning.}
Gemini 3.1 Pro analyzes each video holistically. Rather than merely restating the query, it identifies the central subjects that truly satisfy the event and separates them from auxiliary objects that only describe actions or relations. Distinct physical instances receive separate records, while repeated appearances of the same instance are processed once. If no target satisfies the event, the stage returns an empty result.

For each target, the MLLM selects a frame in which the object is clearly visible, minimally occluded, and distinguishable from similar instances. Long videos are uniformly sampled with an explicit mapping back to original frame indices. The model then generates a description containing category, visible attributes, and spatial relations. Same-category instances receive different appearance and location cues. This converts a cross-frame motion expression into a collection of image-localization tasks with an explicit identity, key frame, and instance-specific description.

\paragraph{SAM3-agent localization and propagation.}
For each instance, Gemini 3.1 Pro interacts with SAM~3 over multiple rounds. It first converts the discriminative description into a concise segmentation phrase, examines the returned candidate masks, and decides whether to accept a result or issue another tool call. Candidate identity, spatial location, and mask completeness are always checked against the full discriminative description, preventing the referred object from drifting when the tool prompt is simplified.

Once a satisfactory pixel-level seed mask is found, it initializes the SAM~3 video tracker on the chosen key frame. The tracker propagates toward both the beginning and end of the sequence. A mask prompt is preferred to a box because it preserves contours and reduces irrelevant regions when targets touch, backgrounds are complex, or several similar objects are present. For plural expressions, every instance is localized and propagated independently before the masks are united frame by frame.

\FloatBarrier

\section{Top Solutions in the MeViSv2-Audio Track}
\label{sec:audio-solutions}

\begin{figure}[t]
  \centering
  \begin{subfigure}[t]{0.49\linewidth}
    \centering
    \includegraphics[width=\linewidth]{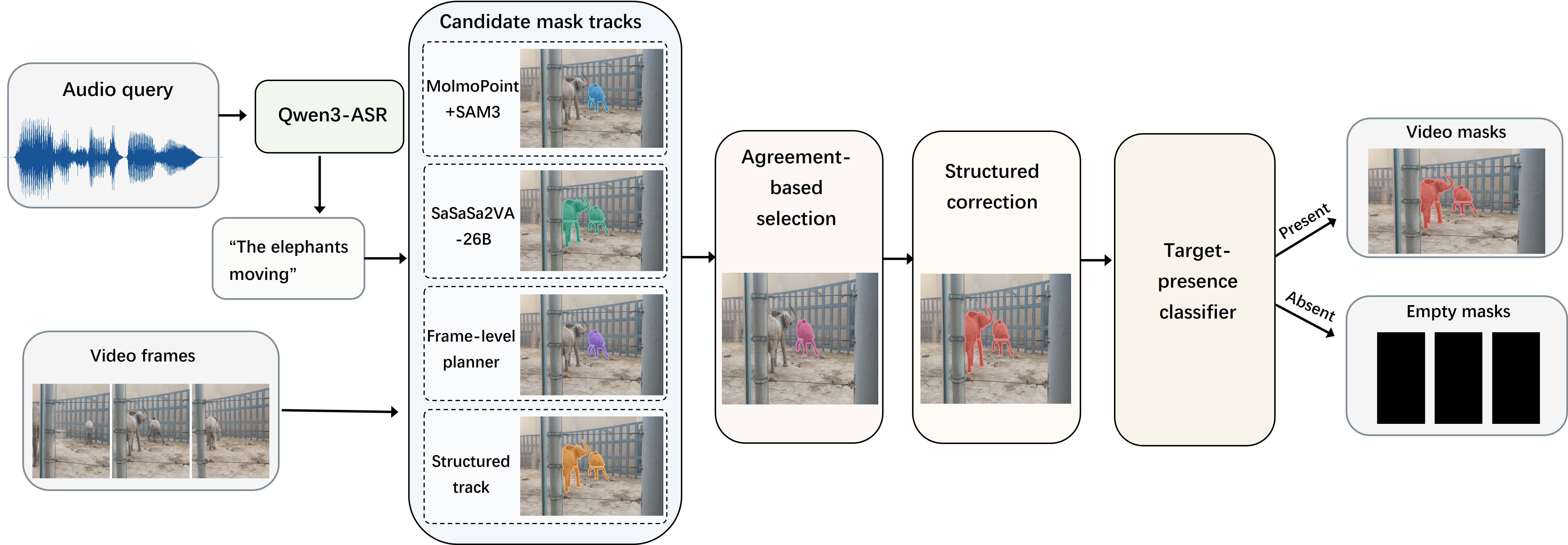}
    \caption{AEXBY selects among complete candidate tracks and applies a target-presence gate.}
    \label{fig:audio1-pipeline}
  \end{subfigure}\hfill
  \begin{subfigure}[t]{0.49\linewidth}
    \centering
    \includegraphics[width=\linewidth]{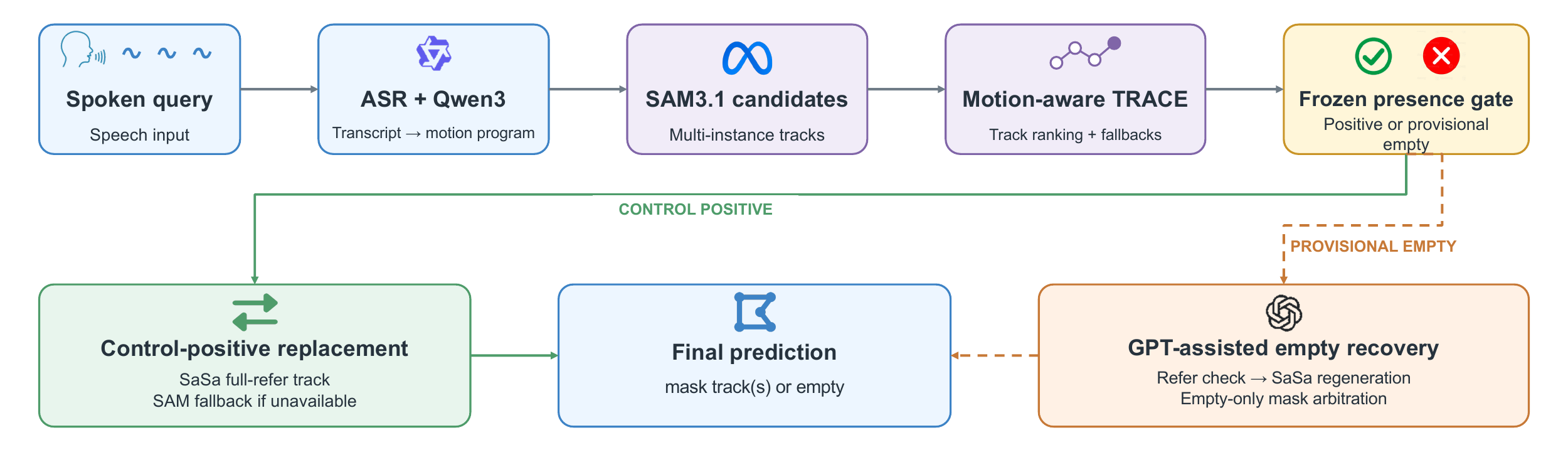}
    \caption{Speech2MaskTrack combines motion-aware ranking with asymmetric recovery.}
    \label{fig:audio2-overview}
  \end{subfigure}
  \caption{Method overviews of the first- and second-place MeViSv2-Audio teams.}
  \label{fig:audio-method-overviews}
\end{figure}

\subsection{1st Place: AEXBY}
\label{sec:audio1}

\teaminfo{Agreement-Based Audio-Visual Segmentation}{Yiwen Ren, Jianing Liu, Yingxin Wang, Kexin Zhang, Licheng Jiao, Lingling Li, Xu Liu}{National 111 Project Base of Intelligent Information Processing}

The winning pipeline first transcribes each audio expression, then generates several candidate mask tracks with different error patterns. A label-free agreement rule chooses one complete track per query, structured corrections handle explicit direction, count, and plural constraints, and a final presence classifier either retains the selected masks or replaces the whole sequence with empty masks.

\paragraph{Audio transcription.}
Qwen3-ASR-1.7B~\cite{shi2026qwen3asr} converts audio $A$ into a text expression $q=\operatorname{ASR}(A)$. Keeping transcription separate from visual reasoning makes the pipeline inspectable and ensures that all candidate generators receive the same query.

\paragraph{Candidate mask tracks.}
The principal candidate uses MolmoPoint-8B~\cite{clark2026molmopoint} to predict object points and timestamps from the video and transcript. These points initialize SAM~3~\cite{carion2026sam}, which propagates masks through the original video. A parallel SAM~2.1 path~\cite{ravi2024sam2} provides an independent consistency signal for a frame-level fallback.

Three complementary candidates are retained. SaSaSa2VA-26B~\cite{niu2025sasasa2va} directly predicts a text-conditioned mask track. A frame-level planner combines point hits, SAM~2.1--SAM~3 agreement, neighboring-frame support, and mask similarity. Finally, a structured route handles explicit horizontal direction, object count, and plural subjects: Qwen3-VL extracts these constraints from sampled frames and SAM~3 creates the corresponding tracks.

\paragraph{Agreement-based selection.}
For $K$ candidate tracks, let $M_i^t$ be candidate $i$ at frame $t$. Their mean temporal overlap is $a(i,j)=T^{-1}\sum_{t=1}^{T}\operatorname{IoU}(M_i^t,M_j^t)$, where two empty masks have overlap one. Each candidate receives $s(i)=(K-1)^{-1}\sum_{j\neq i}a(i,j)$, and the selected index is $i^*=\arg\max_i s(i)$.
The output is the complete track $M_{i^*}$, the medoid of the candidate set under mask overlap. Selection requires no ground-truth labels and does not compare uncalibrated confidence scores from unrelated models.

\paragraph{Structured correction and target presence.}
Agreement may preserve a visually plausible track that violates a query such as ``moving left,'' ``two people,'' or a plural subject. A conservative correction stage activates only for transcripts containing an explicit direction, number, or plural construction. Candidate motion and count are checked, and plural targets are combined by union.

The final presence classifier fuses a visual score, a direct audio-visual score from Qwen2.5-Omni~\cite{xu2025qwen25omni}, and the query's relative rank among expressions for the same video. Raw values, within-video percentiles, standardized scores, distances to video-level extrema, and query count form the input to a balanced logistic regression. If its probability is below a threshold chosen under a high target-recall constraint, all masks are replaced with empty masks. Structured corrections receive a safeguard so that a confidently repaired query is not removed solely by the classifier.

\subsection{2nd Place: StopTheRoll}
\label{sec:audio2}

\teaminfo{Motion-Aware Reasoning from Speech to Mask Tracks}{Jinxing Zhou$^{1}$, Suiyi Zhao$^{2}$, Yanghao Zhou$^{3}$, Ruohao Guo$^{4}$}{$^{1}$Mohamed bin Zayed University of Artificial Intelligence; $^{2}$Anhui University of Science and Technology; $^{3}$National University of Singapore; $^{4}$China Agricultural University}

Speech2MaskTrack delays commitment to a single prediction until it has accumulated evidence over the complete video. After speech recognition and structured query compilation, SAM~3.1 enumerates entity-prompted trajectories. A motion-aware ranker selects a base track, and a frozen lexical presence gate either retains it or produces a provisional empty output. Control-positive predictions may be replaced by a full-expression SaSaSa2VA track, whereas predictions that remain empty enter a separate GPT-assisted recovery route. Recovery may fill an empty result but never overwrite a non-empty mask track.

\paragraph{Speech transcription and structured query.}
Whisper large-v3~\cite{radford2023robust} transcribes the spoken expression while retaining word timing and confidence. A frozen Qwen3 instruction model~\cite{yang2025qwen} compiles the transcript $q_{\mathrm{asr}}$ into a structured motion program $z$. The program contains target category and count, segmentation prompts, reference entities, spatial constraints, and motion atoms describing action, direction, interaction, temporal phase, and target role. Separating targets from reference entities prevents an interacting object from being returned as the final mask.

\paragraph{Candidate generation and motion-aware ranking.}
Target and reference prompts are grounded independently by SAM~3.1~\cite{carion2026sam}. Each candidate $c_k=\{m_{k,t}\}_{t=1}^{T}$ retains masks, boxes, centroids, area, visibility, generator confidence, and prompt role over the complete video. Global camera motion between adjacent frames is estimated with sparse optical flow and robust affine fitting. For observed centroid $p_t$, the residual $\Delta p_t^{\mathrm{res}}=p_{t+1}-\widehat{p}^{\mathrm{cam}}_{t+1}$ isolates motion not explained by camera movement. Complete-trajectory descriptors summarize direction, magnitude, trajectory changes, and early/middle/late activity, along with shape, visibility, relations, and actor--patient compatibility.

Trajectory Ranking with Action-Conditioned Evidence (TRACE) combines a learned compatibility score with expert evidence as $S(c\mid z)=\lambda S_{\mathrm{TRACE}}(c\mid z)+(1-\lambda)S_{\mathrm{expert}}(c\mid z)$. Candidates are sorted by this score; a count-aware rule may retain multiple nonduplicate tracks for plural queries. A frozen lexical presence gate then decides whether the ranked SAM~3.1 base prediction is control-positive or provisionally empty.

\paragraph{Replacement and recovery.}
For a control-positive query, a usable SaSaSa2VA track~\cite{niu2025sasasa2va} directly replaces the SAM~3.1 mask; the two backends are neither score-compared nor mask-averaged. If SaSaSa2VA has no usable track, the highest-ranked SAM~3.1 result remains. A control-absent decision bypasses this replacement.

Only outputs that remain empty enter recovery. GPT first adjudicates the transcript against a chronological raw-video storyboard and normalizes the query only when visual evidence supports a repair. SaSaSa2VA regenerates a candidate only when GPT predicts target presence, a full constraint match, a non-empty answer entity, and sufficient confidence. A second GPT call acts as mask arbiter: it compares raw and mask-overlay storyboards and verifies identity, class, count, attributes, action, semantic role, relation, temporal evidence, and mask geometry. GPT never produces mask pixels. Formally, with gated base $P_i$, main replacement $S_i$, and recovery $R_i$,
\begin{equation}
  B_i=\begin{cases}S_i,&P_i\neq\varnothing\land S_i\neq\varnothing,\\P_i,&\text{otherwise},\end{cases}
  \quad
  M_i=\begin{cases}R_i,&B_i=\varnothing\land\operatorname{accept}(R_i),\\B_i,&\text{otherwise}.\end{cases}
  \label{eq:audio2-asymmetric-merge}
\end{equation}
The asymmetry protects trusted non-empty predictions while allowing carefully verified recall recovery.

\subsection{3rd Place: Agent-VOS}
\label{sec:audio3}

\teaminfo{MLLM-Assisted Audio VOS}{Liangtao Shi$^{1}$, Jinxia Xie$^{2}$, Xiantao Hu$^{2}$, Ting Liu$^{3}$}{$^{1}$Hefei University of Technology; $^{2}$Nanjing University of Science and Technology; $^{3}$Hunan Police College}

Agent-VOS is a training-free audio-referring VOS pipeline in which MLLMs perform speech and video-language understanding while SAM-based models perform dense segmentation and tracking. The method contains four stages: audio-to-text conversion, joint video-text analysis, text-based segmentation with mask-guided tracking, and mask-text consistency verification.

\paragraph{Audio-to-text conversion.}
Qwen3-ASR-1.7B~\cite{shi2026qwen3asr} transcribes audio $A$ into text $q=\Phi_{\mathrm{ASR}}(A)$. This conversion allows the pipeline to reuse text-conditioned video understanding and segmentation models without task-specific training. Because a transcript may be too coarse for multiple targets, fine-grained attributes, or complex temporal cues, it is subsequently refined using the video.

\paragraph{Joint video-text analysis.}
Gemini 3 Flash Preview jointly analyzes $q$ and video $V$. It determines the number of referred instances and produces for every target a tuple $o_i=(d_i,k_i)$, where $d_i$ is an instance-specific description and $k_i$ is a representative key frame. The description encodes visual and contextual cues that distinguish the instance; the selected frame provides a reliable initialization point for tracking.

\paragraph{Segmentation and mask-guided tracking.}
MomentSeg~\cite{dai2025momentseg} predicts the coarse sequence $\widetilde{M}_i=\{\widetilde{m}_{i,t}\}_{t=1}^{T}=\Phi_{\mathrm{MomentSeg}}(V,d_i)$. Its mask at the selected key frame, $\widetilde{m}_{i,k_i}$, initializes DAM4SAM~\cite{videnovic2025distractor}, which refines and propagates the target bidirectionally as $\widehat{M}_i=\Phi_{\mathrm{DAM4SAM}}(V,\widetilde{m}_{i,k_i},k_i)$.
Multiple instances are processed independently and their mask sequences are aggregated for the final prediction.

\paragraph{Mask-text consistency verification.}
Propagation can drift toward visually similar distractors or fail through occlusion. To detect such errors, all instance masks are merged and overlaid on the video. Gemini 3 Flash Preview receives the overlay together with the transcript and verifies whether the highlighted object remains semantically consistent with the query. An inconsistent prediction is replaced by an empty mask; otherwise it is retained. The verifier therefore provides a final semantic gate without modifying mask pixels.

\FloatBarrier

\section{Conclusion and Discussion}
\label{sec:overall-conclusion}

Across all three tracks, leading solutions use foundation models such as SAM~3 as shared mask-generation and propagation engines, while task-specific modules decide how to prompt them. MOSEv2 methods preserve identity through target routing, memory, and corrective re-initialization under occlusion, reappearance, and same-category interference.

Text and audio systems treat query understanding and target existence as explicit reasoning problems. They decompose expressions, verify tracks, distinguish absence from invisibility, and arbitrate masks; audio entries also separate transcription from grounding. Agentic loops inspect overlays and revise uncertain predictions. Future work should unify semantic reasoning, memory, and propagation while reducing multi-model cost.

\section*{Acknowledgements}
This work was supported by the National Natural Science Foundation of China (NSFC) under Grant No.~62472104, the Science and Technology Commission of Shanghai Municipality under Grant No.~25511103600, and Shanghai Pujiang Program 2025PJA201.

\bibliographystyle{splncs04}
\bibliography{main}

\end{document}